\documentclass[10pt,twocolumn,letterpaper]{article}
\usepackage{times}
\usepackage{epsfig}
\usepackage{graphicx}
\usepackage{amsmath}
\usepackage{amssymb}
\usepackage{subfigure}
\usepackage{comment}
\usepackage{booktabs}
\usepackage[pagebackref=true,breaklinks=true,letterpaper=true,colorlinks,bookmarks=false]{hyperref}

\begin{document}

\title{Learning to Localize in Unseen Scenes with Relative Pose Regressors}
\author{Ofer Idan, Yoli	Shavit, Yosi Keller\thanks{%
yosi.keller@gmail.com} \\
Bar Ilan University}
\date{}
\maketitle

\begin{abstract}
Relative pose regressors (RPRs) localize a camera by estimating its relative
translation and rotation to a pose-labelled reference. Unlike scene
coordinate regression and absolute pose regression methods, which learn
absolute scene parameters, RPRs can (theoretically) localize in unseen
environments, since they only learn the residual pose between camera pairs.
In practice, however, the performance of RPRs is significantly degraded in
unseen scenes. In this work, we propose to aggregate paired feature maps
into latent codes, instead of operating on global image descriptors, in
order to improve the generalization of RPRs. We implement aggregation with
concatenation, projection, and attention operations (Transformer Encoders)
and learn to regress the relative pose parameters from the resulting latent
codes. We further make use of a recently proposed continuous representation
of rotation matrices, which alleviates the limitations of the commonly used
quaternions. Compared to state-of-the-art RPRs, our model is shown to
localize significantly better in unseen environments, across both indoor and
outdoor benchmarks, while maintaining competitive performance in seen
scenes. We validate our findings and architecture design through multiple
ablations. Our code and pretrained models is publicly available\footnote{%
\url{https://github.com/yolish/relformer}}.
\end{abstract}



\begin{figure}[tbh!]
\begin{center}
\subfigure[Relative Pose Regression
Schemes]{\includegraphics[scale=0.29]{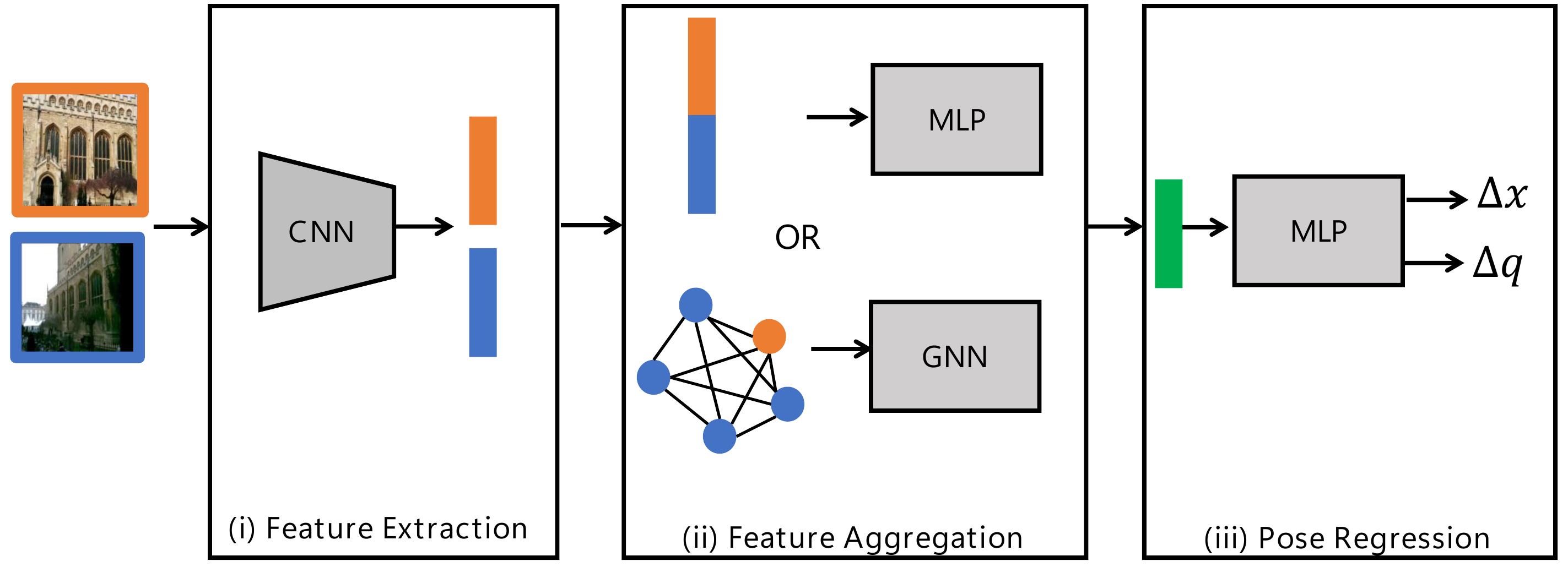}}
\subfigure[Relative
Pose Regression with Improved Generalization (Proposed)] {\includegraphics[scale=0.29]{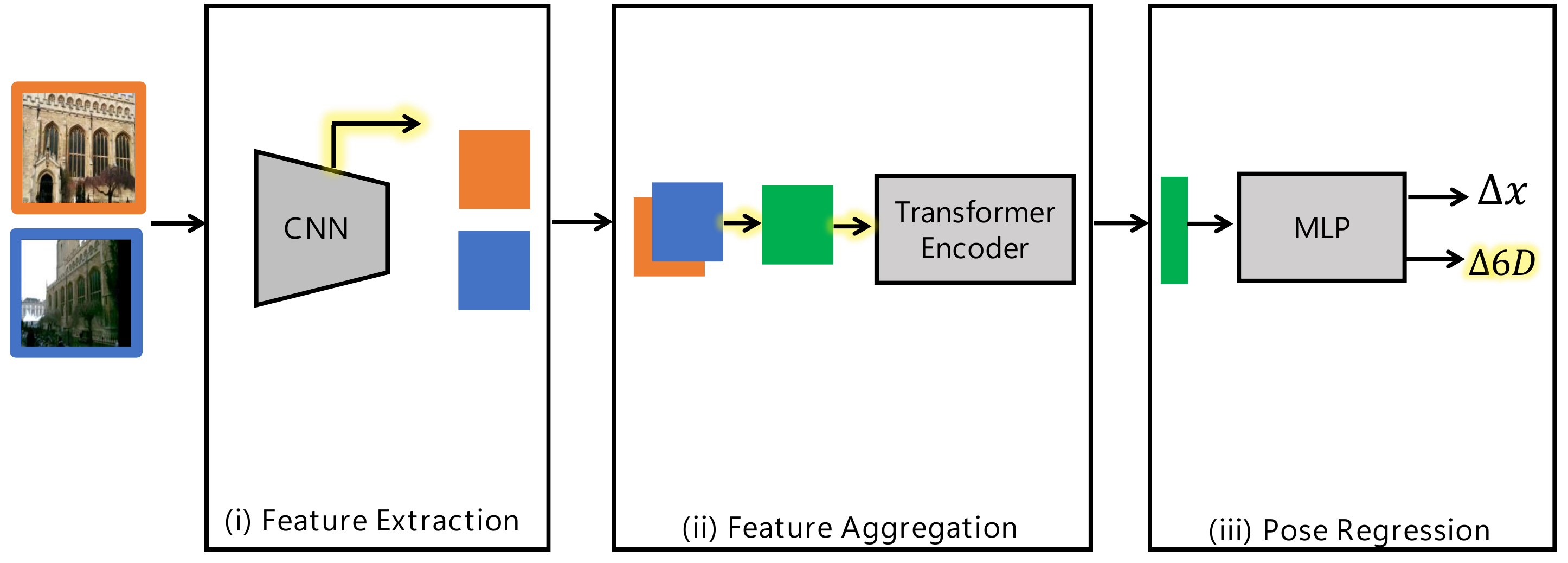}}
\end{center}
\caption{Schemes for Relative Pose Regression.}
\label{fig:rpr}
\end{figure}

\section{Introduction}

Image-based localization is an essential step in various computer vision
applications, ranging from autonomous navigation to positioning avatars in augmented or virtual reality. Some methods aim to tackle this problem
using the query image alone. Scene coordinate regression (SCR) methods \cite%
{DSAC,DSAC++} predict the 3D scene coordinates from the 2D pixels of the
input image and then apply Perspective-n-Point (PnP) and RANSAC \cite%
{fischler1981random} to predict the camera pose. Absolute Pose Regression
(APR) methods \cite%
{kendall2015posenet,kendall2017geometric,shavit2021learning} directly
regress the pose parameters from the query image. While offering fast and
standalone solutions, both SCR and APR are confined by the absolute
coordinates of the scenes and cannot generalize to unseen environments.

Structure-based localization pipelines (SLPs) trade memory and speed for
generalization, while achieving state-of-the-art localization accuracy in
challenging conditions \cite{taira2018inloc,sarlin2019coarse, sarlin21pixloc}%
. Extracting and matching global image descriptors, local features, and
depth information allows establishing 2D-3D correspondences between query
and visually similar reference images in order to estimate the camera pose
using PnP and RANSAC. Since image retrieval, feature extraction, and feature
matching are not scene-specific tasks, such pipelines can localize in new
environments, without model retraining.

Relative Pose Regressors (RPRs) predict the parameters of the relative pose
between a pair of cameras. Most contemporary RPRs first extract global
features from query and reference images, using a convolutional backbone
(Fig. \ref{fig:rpr}a-i). The query and reference features are then joined
and aggregated together (Fig. \ref{fig:rpr}a-ii). Aggregation is typically
performed through concatenation and one or more Fully Connected (FC) layers
\cite{nn-net, ding2019camnet,balntas2018relocnet}, or more recently, with
Graphical Neural Networks (GNN) \cite{turkoglu2021visual}. The resulting
latent aggregation is passed to a multi-layer perceptron (MLP) which
regresses the camera pose, typically represented with a 3D translation
vector, $\Delta x$, and the quaternion parameterization of the relative
rotation, $\Delta q$ (Fig. \ref{fig:rpr}a-iii). The pose of the query image
is then computed by matrix inversion and multiplication. Some RPRs estimate
instead the essential matrix and then apply RANSAC to estimate the pose of
the query image \cite{zhou2020learn}.

Since RPRs learn to predict residuals, they can theoretically localize in
scenes not seen during training ('unseen'). In practice, however, current
methods present a significant degradation in performance in unseen
environments \cite{zhou2020learn,turkoglu2021visual}. In this work, we study
the generalization of RPRs. We consider the three main operations performed
by relative pose regression methods: feature extraction, joint feature
aggregation and pose regression. We find that operating on feature maps,
available from intermediate activations, is key to localizing in unseen
scenes (Fig \ref{fig:rpr}b-i). Thus, we propose Relformer, an RPR model that
concatenates, projects, and aggregates feature maps into latent codes using
Transformer Encoders (Fig \ref{fig:rpr}b-ii). We further adopt a recently
proposed 6D representation for rotation matrices \cite{zhou2019continuity}
for our learning objective, shown to be advantageous for regression tasks,
compared to quaternions (Fig \ref{fig:rpr}b-iii). We evaluate our method on
indoor and outdoor benchmarks when localizing in 'seen' and 'unseen'
environments. Relformer is shown to improve generalization to unseen scenes
by a significant margin, compared to the current state-of-the-art RPRs,
while achieving competitive performance on seen scenes.

In summary, our contributions are as follows:

\begin{itemize}
\item We show that operating on feature maps is key to improving the
localization of RPRs in unseen scenes.

\item We propose Relformer, a novel RPR architecture, which concatenates,
projects, and aggregates feature maps with Transformer Encoders to predict
the relative pose parameters.

\item Our model surpasses current state-of-the-art RPRs in unseen
environments, while maintaining competitive results on seen scenes.
\end{itemize}

\section{Related Work}

\subsection{Visual Localization}

Methods for image-based camera localization typically estimate the camera
pose parameters in one of three ways: (1) with image retrieval (IR), (2) by
establishing correspondences between 2D pixels and 3D scene coordinates and
applying PnP-RANSAC (assuming known camera intrinsics) (3) by regressing
explicit camera parameters from which the camera pose can be recovered
(e.g., the camera pose parameters or a relative camera pose). IR methods \cite%
{denseVLAD,arandjelovic2016netvlad,dusmanu2019d2,radenovic2018fine,hausler2021patch}
can localize images by taking the pose of the closest neighbor as the pose
of the query image, or by interpolating the poses of several neighbors \cite%
{benchmarking_ir3DV2020,sattler2019understanding}. Methods in the second class, often referred to as 'structure-based', include
SLPs and SCR approaches. SLPs \cite{taira2018inloc, sarlin2019coarse,
sarlin21pixloc} find 2D-2D matches between the local features of the query
and the reference images, obtained with IR, and then use depth labels to
establish 2D-3D pairs. Instead of storing heavy-memory global and local
image features, SCR methods \cite{DSAC,DSAC++} directly regress the 3D world
coordinates from the 2D pixels of the query image. Both SLPs and SCR achieve
state-of-the-art localization accuracy. SLPs have high storage and runtime
requirements, but can operate in challenging conditions and generalize to
unseen scenes. SCR methods, in turn, are fast and estimate the camera pose
using the query image alone. However, they can only localize in scenes seen
during training and need to be trained per scene. The last family of methods
proposes to directly estimate camera parameters from images. APRs are
typically trained per scene, encoding images with a convolutional backbone
and then regressing the camera pose parameters with a multi-layer perceptron
(MLP) \cite%
{kendall2015posenet,kendall2016modelling,kendall2017geometric,melekhov2017image,naseer2017deep,wu2017delving,shavitferensirpnet}%
. This scheme was recently extended to learn multiple scenes with a single
model using Transformers \cite{ShavitFerensIccv21} or by indexing
scene-specific weights \cite{blanton2020extending}. Pose encoding was also
proposed as a means for introducing scene priors and improving performance
\cite{shavit2022camera}. \newline
Relative pose regression methods regain generalization by learning the
relative pose parameters instead of the absolute ones. Laskar et al.\cite%
{nn-net}, were the first to suggest this approach using a Siamese ResNet34
backbone to encode a pair of images into global image descriptors and then
regressed the relative pose parameters from their concatenation with an MLP.
At test time, the model was applied to encode the query image and to regress
the relative pose with respect to an image fetched from a reference database
based on the distance between the query and reference encoding. The query
camera pose was then estimated from the known pose of the reference image
and the predicted relative pose, through matrix inversion and
multiplication. The authors estimated pose candidates using several closest
neighbors and then predicted the final pose through averaging algorithms.
This approach was able to localize scenes not included in the model's
training set, albeit with a significant degradation in accuracy.
Modifications to loss and training procedure were later proposed and
achieved improved performance in seen environments \cite%
{ding2019camnet,balntas2018relocnet}, matching or surpassing the accuracy of
APRs. More recently, GNNs were suggested for exchanging information between
multiple image encodings by learning and pruning the complete graph formed
from the query and multiple neighbor images \cite{turkoglu2021visual}.
Instead of directly learning the relative pose parameters, Zhou et al.
suggested learning the essential matrix from paired images \cite%
{zhou2020learn} to recover the query camera pose.\newline
Other methods employed relative pose regression for estimating the motion
with respect to a set of predicted anchor points \cite{saha2018improved} or
to iteratively optimize an initial pose matrix by maximizing the similarity
between the query image and the image synthesized with a neural radiance
field \cite{yen2020inerf} (pre-trained on the scene of interest). However,
these approaches are scene-specific and, as opposed to RPRs, cannot localize
in an unseen scene. \newline
Compared to APRs, RPRs are slower and require the storage of image
encodings. However, they achieve similar accuracy and can localize in unseen
scenes. Compared to structured-based methods, RPRs (and regression
approaches in general) present inferior localization accuracy. However, they
are faster and have lower storage requirements than SLPs, while offering the
ability to generalize to unseen scenes, which is lacking in SCR approaches.
In practice, the performance of RPRs in unseen scenes significantly degrades
compared to seen environments. In this work, we propose a method for
improving the generalization of RPRs while maintaining performance in seen
scenes.

\subsection{Regressing Rotations with Deep Learning}

RPRs typically regress the camera pose as a tuple $\mathbf{p}:<\mathbf{x},%
\mathbf{q}>$, where $\mathbf{x}\in \mathbb{R}^{3}$ gives the translation
vector, and $\mathbf{q}\in \mathbb{R}^{4}$ represents the unit quaternion
parameterization of the camera rotation. The $\mathbb{L}_{2}$ or $\mathbb{L}%
_{1}$ losses are commonly used for optimizing the prediction of $\mathbf{x}$
and $\mathbf{q}$, with respect to their ground truth values, and are joined
together through manual \cite{kendall2015posenet,shavitferensirpnet} or
learned weights\cite{kendall2017geometric}. Brahmbhatt et al.\cite%
{brahmbhatt2018geometry} further proposed to use the log (unit) quaternion
to improve performance. However, quaternions are not an optimal
parameterization of rotation matrices. First, they suffer from an antipodal
problem, where $\mathbf{q}$ and $-\mathbf{q}$ represent the same rotation
matrix. Furthermore, it was recently shown that representations of 3D
rotations in $\le4$D Euclidean real spaces are discontinuous\cite%
{zhou2019continuity}. Discontinuous representations, such as quaternions and
Euler angles, were empirically shown to be inferior for regressing the
rotation parameters, compared to continuous representations \cite%
{zhou2019continuity}. Consequently, different continuous representations for
rotation regression were proposed \cite%
{zhou2019continuity,levinson2020analysis,peretroukhin2020smooth,chen2022projective}%
. Zhou et al. \cite{zhou2019continuity} proposed learning a 6D vector,
composed of the two first column vectors of the rotation matrix, and to
recover the rotation matrix through a Gram-Schmidt orthogonalization
process. Levinson et al. \cite{levinson2020analysis} instead suggested a 9D
parameterization, which is based on the singular value decomposition (SVD)
orthogonalization. In this work, we choose the 6D representation for our
learning objective (see the ablations in Tables \ref{tb:ablation-seen},\ref%
{tb:ablation-unseen},\ref{tb:ablation-rot}).

\subsection{Transformers in Visual Localization}

Vaswani et al. \cite{AttentionIsAllYouNeed} introduced transformers to
encode and aggregate sequences with attention layers \cite%
{DBLP:journals/corr/BahdanauCB14}, in order to solve sequence-to-sequence
problems in Natural Language Processing (NLP). Attention layers update each
element in a sequence through weighted aggregation of all elements based on
pairwise correlations. The attention mechanism is also related to bilinear
pooling\cite{bilipool}, which has been shown to improve image recognition at
fine-grained levels. \newline
Transformers were shown to outperform sequential neural models, such as RNNs
and LSTMs in encoding long sequences, and were successfully applied for
various tasks of natural language processing (NLP) and computer vision \cite%
{bert}. In the context of localization, Transformers were shown to achieve
state-of-the-art performance for image retrieval \cite{el2021training},
object detection\cite{DETR}, feature detection, extraction and matching \cite%
{jiang2021cotr,sun2021loftr} and for multiscene absolute pose regression
\cite{shavit2021learning}.\newline
Encouraged by a recent observation that hybrid architectures of CNN and
Transformers outperform CNN- and Transformer-Only architectures \cite%
{xiao2021early} and the success of Transformers in learning camera pose
regression\cite{shavit2021learning}, we propose a novel CNN-Transformer RPR
architecture which aggregates paired feature maps into latent
representations of relative pose parameters.

\section{Method}

\begin{figure*}[tbh!]
\begin{center}
\includegraphics[scale=0.55]{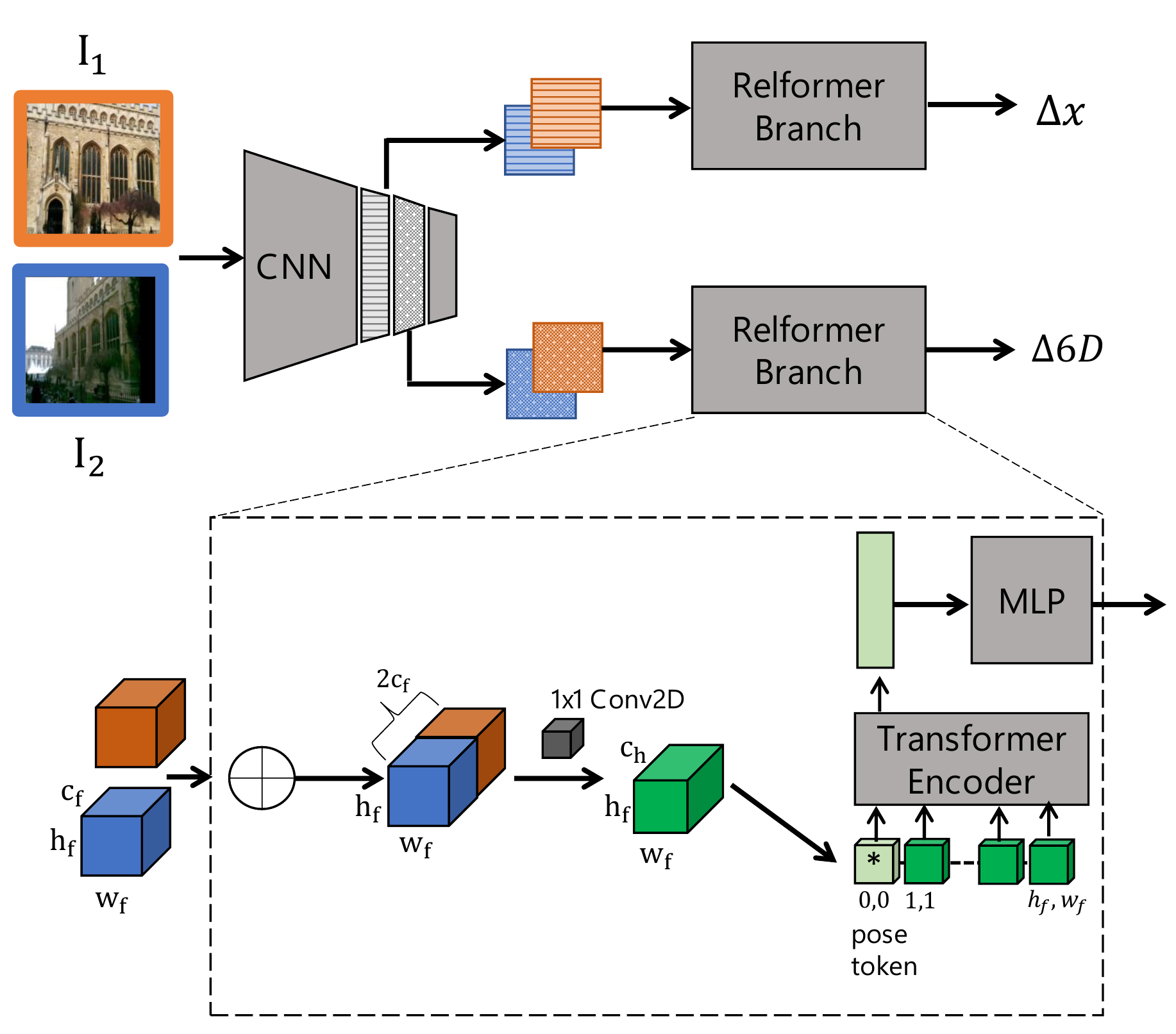}
\end{center}
\caption{The architecture of our proposed model (Relformer).}
\label{fig:arch}
\vspace{-6mm}
\end{figure*}
We implement our method using the hybrid CNN-Transformer architecture, we call
Relformer, which is depicted in Fig. \ref{fig:arch}. We apply a shared
convolutional backbone to extract feature maps from image pairs and apply
two separate branches to predict the translation and rotation parameters,
respectively. Each branch is implemented with a Transformer Encoder and an
MLP head. The Transformer Encoders aggregate paired feature maps, extracted
from query and reference images, into a latent representation of the pose
parameters. Each MLP head regresses its target ($\Delta _{x}$ or $\Delta 6D$%
) from the respective latent representation.

\subsection{Feature Extraction}

Given a pair of images $\mathbf{I_1,I_2}\in \mathbb{R}^{H\times W\times C}$,
we apply a Siamese convolutional backbone to extract feature maps. Similarly
to \cite{ShavitFerensIccv21}, we extract a feature map per regression
target: $\mathbf{F}^{\mathbf{i}}_{\mathbf{trans}}$ and $\mathbf{F}^{\mathbf{i%
}}_{\mathbf{rot}}$, for learning translations and rotations, respectively
(for image $i$=$1,2$)

\subsection{Paired Feature Aggregation}

The extracted feature maps capture visual information from each image. Given
the feature maps $\mathbf{F}_{\mathbf{t}}^{\mathbf{1}}$,$\mathbf{F}_{\mathbf{%
t}}^{\mathbf{2}}$ $\in \mathbb{R}^{H_{f}\times W_{f}\times C_{f}}$,
extracted for learning the regression target $\mathbf{t}$ from the pair $%
\mathbf{I_{1}}$,$\mathbf{I_{2}}$, we compute a \textit{paired} feature map $%
\mathbf{F}_{\mathbf{t}}\in \mathbb{R}^{H_{f}\times W_{f}\times C_{h}}$ by
applying channelwise concatenation and linear projection with a $1\times 1$
convolution (projecting to dimension $C_{h}$). We also assign an encoding to
each position in our paired feature map. Specifically, we learn a 2D
position encoding as in \cite{DETR}, by defining the embedding vectors $%
\mathbf{E}_{x}\in \mathbb{R}^{\left( W_{f}+1\right) \times C_{h}/2}$ and $%
\mathbf{E}_{y}\in \mathbb{R}^{\left( H_{f}+1\right) \times C_{h}/2}$, such
that a position $\left( i,j\right) ,$ $i\in 1..H_{f}$, $j\in 1..W_{f}$, is
encoded by the concatenation of the corresponding embedding vectors:%
\begin{equation}
\mathbf{E}_{pos}^{i,j}=%
\begin{bmatrix}
\mathbf{E}_{x}^{j} \\
\mathbf{E}_{y}^{i}%
\end{bmatrix}%
\in \mathbb{R}^{C_{h}}.
\end{equation}%
The paired feature map and its position encoding are then flattened into a
sequence $\widehat{\mathbf{F_{t}}}\in \mathbb{R}^{{H_{f}}\cdot W_{f}\times
C_{h}}$. \newline
In order to aggregate our sequence into a latent vector representation, we
follow recent methods for sequence classification \cite{bert}, and append a
learned task token $\mathbf{r_{t}}\in \mathbb{R}^{C_{h}}$ to the flattened
feature map $\widehat{\mathbf{F_{t}}}$. The pose token is assigned to the
position $(0,0)$ and the embedding vectors $(\mathbf{E}_{x}^{0},\mathbf{E}%
_{y}^{0})$ and is appended to the position encoding sequence. Therefore, the input to the Transformer Encoder learning task $\mathbf{t}$ is given by: $[\mathbf{r_{t}},\widehat{\mathbf{F_{t}}}]\in \mathbb{R}^{({H_{f}}\cdot W_{f}+1)\times C_{h}}$.

We use the Transformer-Encoder architecture implementation of \cite{DETR},
which includes $n$ identical Encoder layers. Each layer consists of a
multi-head self-attention (MHA) module and an MLP with two layers and gelu
non-linearity. The learned position encoding is added to the input before
applying each Transformer Encoder layer. A LayerNorm \cite{ba2016layer} is
applied before each module (MHA/MLP) and the input is added to the output
with residual connections \cite{AttentionIsAllYouNeed} and dropout. We take
the output of the encoder at the position of the special token: $\mathbf{%
r_{t}}^{\prime}\in \mathbb{R}^{C_{h}}$, which aggregates the information
from the entire paired sequence.

\subsection{Pose Regression}

We apply an MLP with a single hidden layer and gelu non-linearity to regress
the target vector of task $\mathbf{t}$ from $\mathbf{r_{t}}^{\prime}$.
Specifically, we regress $\mathbf{\Delta x}\in \mathbb{R}^{3}$, the
translation vector, and $\mathbf{\Delta 6D}\in \mathbb{R}^{6}$, the 6D
parameterization of the rotation matrix $\mathbf{\Delta R}\in \mathbb{R}%
^{3\times3}$. Given the known pose of $\mathbf{I_1}: <\mathbf{x_{1}},
\mathbf{R_{1}}>$, we can recover the pose of $\mathbf{I_2}$ using $\mathbf{%
\Delta x}$ and $\mathbf{\Delta R}$:
\begin{equation}
\mathbf{x_{2}} = \mathbf{x_{1}} + \mathbf{\Delta x}, \mathbf{R_{2}} =
\mathbf{R_{1}}\mathbf{\Delta R},  \label{equ:pose}
\end{equation}

\subsection{Training Loss for Relative Camera Pose}

We optimize our model with the pose loss suggested by \cite%
{kendall2017geometric} for learning the relative weights of the translation
and rotation losses $L_{\mathbf{\Delta x}}$ and $L_{\mathbf{\Delta 6D}}$:
\begin{equation}
L_{\mathbf{p}}=L_{\mathbf{\Delta x}}\exp (-s_{\mathbf{\Delta x}})+s_{\mathbf{%
\Delta x}}+L_{\mathbf{\Delta 6d}}\exp (-s_{\mathbf{\Delta 6d}})+s_{\mathbf{%
\Delta 6d}},  \label{equ:learnable pose loss}
\end{equation}%
where $s_{\mathbf{\Delta x}}$ and $s_{\mathbf{\Delta 6D}}$ are learned
parameters. $L_{\mathbf{\Delta x}}$ and $L_{\mathbf{\Delta 6D}}$ are given
by the $\mathbb{L}_{1}$ norm of the difference between the predicted and
ground truth values. We recover the rotation matrix from the predicted 6D
representation using Gram-Schmidt orthogonalization \cite%
{zhou2019continuity}.

\begin{table*}[tbh]
\caption{\textbf{Localizing with RPRs in indoor previously seen scenes
(7Scenes).} We report the median position and orientation errors (in
meters/degrees, respectively) for each scene and the average of the medians in
all scenes. The best and second-best results are highlighted (in
bold/underline, respectively).}
\label{tb:7scenes-seen}\centering
\begin{tabular}{l|ccccccc|c}
\toprule
\textbf{Method} & \textbf{Chess} & \textbf{Fire} & \textbf{Heads} & \textbf{%
Office} & \textbf{Pumpkin} & \textbf{Kitchen} & \textbf{Stairs} & \textbf{%
Avg.} \\ \midrule
{NN-Net}\cite{nn-net} & 0.13/{6.50} & {0.26 }/{12.7} & \underline{0.14}/{12.3%
} & 0.21/{7.40} & {0.24}/{6.40} & {0.24}/{8.00} & {0.27}/{11.8} & 0.21/9.30
\\
{RelocNet}\cite{balntas2018relocnet} & 0.12/{4.10} & {0.26 }/{10.4} &
\underline{0.14}/{10.5} & 0.18/{5.30} & {0.26}/{4.20} & {0.23}/{5.10} & {0.28%
}/{7.50} & 0.21/6.70 \\
{CamNet}\cite{balntas2018relocnet} & --/-- & --/-- & --/-- & --/-- & --/-- &
--/-- & --/-- & \textbf{0.05}, \textbf{1.80} \\
{EssNet \cite{zhou2020learn}} & 0.13/{5.10} & {0.27}/{10.1} & {0.15}/%
\underline{9.90} & 0.21/{6.90} & {0.22}/{6.1} & {0.23}/{6.90} & {0.32}/{11.2}
& 0.22/8.00 \\
{NC-EssNet \cite{zhou2020learn}} & 0.12/{5.60} & {0.26 }/{9.60} & \underline{%
0.14}/{10.7} & 0.20/{6.70} & {0.22}/{5.70} & {0.22}/{6.30} & {0.31}/{7.90} &
0.21/7.50 \\
RelPoseGNN \cite{turkoglu2021visual} & \textbf{0.08}/\textbf{2.70} & \textbf{%
0.21 }/\textbf{7.50} & \textbf{0.13}/\textbf{8.70} & \textbf{0.15}/\textbf{%
4.10} & \textbf{0.15}/\textbf{3.50} & \textbf{0.19}/\textbf{3.70} & \textbf{%
0.22}/\underline{6.50} & \underline{0.16}/\underline{5.20} \\
\textbf{Relformer (Ours)} & \underline{0.11}/\underline{4.01} & \underline{%
0.23}/\underline{8.57} & 0.17/10.9 & \underline{0.16}/\underline{4.92} &
\textbf{0.15}/\underline{4.15} & \textbf{0.19}/\underline{4.89} & \underline{%
0.24}/\textbf{6.46} & 0.18/6.27 \\ \bottomrule
\end{tabular}%
\end{table*}
\begin{table*}[tbh]
\caption{\textbf{Localizing with RPRs in unseen scenes (7Scenes).} The
methods were trained with training sets of six scenes and evaluated on the
remaining scene. We report the median position and orientation errors (in
meters/degrees, respectively) for each scene and the average of medians
across all scenes. Results for methods marked with '*' were reported and
reproduced by \protect\cite{turkoglu2021visual}. The best and second-best
results are highlighted (with bold/underline, respectively). }
\label{tb:7scenes-unseen}\centering
\begin{tabular}{l|ccccccc|c}
\toprule
\textbf{Method} & \textbf{Chess} & \textbf{Fire} & \textbf{Heads} & \textbf{%
Office} & \textbf{Pumpkin} & \textbf{Kitchen} & \textbf{Stairs} & \textbf{%
Avg.} \\ \midrule
NN-Net\cite{nn-net}* & 0.33/\textbf{11.5} & 11.68 /\textbf{13.8} & {0.30}/{%
15.5} & 1.34/\textbf{10.9} & \underline{0.41}/{12.8} & {1.68}/{12.9} & {0.44}%
/\underline{13.6} & 2.31/13.0 \\
{EssNet\cite{zhou2020learn}}* & 0.73/37.6 & 0.89/67.6 & 0.62/28.5 & 0.84/36.3
& 1.06/33.3 & 0.91/36.1 & 1.19/42.1 & 0.89/40.2 \\
NC-EssNet\cite{zhou2020learn}* & 0.62/24.2 & 0.75/23.7 & 0.44/25.6 &
0.88/28.0 & 1.02/24.5 & 0.77/20.8 & 1.25/36.5 & 0.82/26.2 \\
RelPoseGNN\cite{turkoglu2021visual} & \underline{0.29}/12.8 & \underline{0.45%
}/15.7 & \textbf{0.19}/\underline{14.7} & \underline{0.42}/12.5 & 0.44/%
\underline{11.7} & \underline{0.42}/\underline{12.4} & \textbf{0.35}/15.5 &
\underline{0.36}/\underline{13.6} \\
\textbf{Relformer (Ours)} & \textbf{0.27}/\textbf{8.49} & \textbf{0.34}/%
\textbf{10.4} & \underline{0.28}/\textbf{12.1} & \textbf{0.32}/\textbf{8.24}
& \textbf{0.40}/\textbf{9.06} & \textbf{0.32}/\textbf{9.23} & \underline{0.42%
}/\textbf{10.5} & \textbf{0.34}/\textbf{9.72} \\ \bottomrule
\end{tabular}%
\vspace{-6mm}
\end{table*}
\begin{table}[tbh]
\caption{\textbf{Localizing with RPRs in outdoor previously seen and unseen
environments (Cambridge Landmarks).} We report the average of the median
position and orientation errors (in meters/degrees, respectively) when
training on the CambridgeLandmarks dataset (Seen) and when training on the
7Scenes dataset (Unseen). In both cases, the models are evaluated on four
scenes from the Cambridge Landmarks dataset. The best and second-best
results are highlighted (in bold/underline, respectively).}
\label{tb:cambridge}\centering
\begin{tabular}{l|cc}
\toprule
\textbf{Method} & \textbf{Seen} & \textbf{Unseen} \\ \midrule
{EssNet \cite{zhou2020learn}} & 1.07/3.42 & 10.4/85.8 \\
{NC-EssNet \cite{zhou2020learn}} & \underline{0.84}/\underline{2.83} &
\underline{7.98}/\underline{24.4} \\
RelPoseGNN \cite{turkoglu2021visual} & \textbf{0.91}/\textbf{2.3} & --/-- \\
\textbf{Relformer (Ours)} & 1.33/3.52 & \textbf{3.35}/\textbf{10.6} \\ \bottomrule
\end{tabular}%
\vspace{-6mm}
\end{table}
\begin{table*}[tbh]
\caption{Cross-Comparison of different localization methods in previously
seen and unseen environments. We consider representative methods from five
localization families, which present different runtime and storage
requirements (indicated as low/intermediate/high with '+'/'++'/'+++',
respectively) and vary in their ability to learn multiple scenes with a
single model (Multi-Scene) and to generalize to unseen scenes
(Generalization), which we indicate with a $\checkmark $ sign. For each
method, we provide the reported average of median position and orientation
errors (in meters/degrees, respectively) for seen and unseen localization on
the 7Scenes dataset. For RPR methods, we report seen and unseen statistics
in the same manner as in Tables \protect\ref{tb:7scenes-seen} and \protect
\ref{tb:7scenes-unseen}. We report the performance of IR and SLP methods,
which were trained on different large datasets, in the Unseen category.
Results marked with * were reported and reproduced by \protect\cite%
{turkoglu2021visual}.The 'F' character in the Unseen column indicates that
the method fails to localize in unseen scenes.}
\label{tb:cross}\centering
\begin{tabular}{lccccc||cc}
\toprule
\textbf{Method} & \textbf{Family} & \textbf{Storage} & \textbf{Runtime} &
\textbf{Multi-Scene} & \textbf{Generalization} & \textbf{Seen} & \textbf{%
Unseen} \\ \midrule
DenseVLAD\cite{denseVLAD} & IR & ++ & ++ & \checkmark & \checkmark & -- &
0.26/12.5 \\
DenseVLAD+inter.\cite{sattler2019understanding} & IR & ++ & ++ & \checkmark
& \checkmark & -- & 0.24/11.7 \\ \midrule
InLoc\cite{taira2018inloc} & SLP & +++ & +++ & \checkmark & \checkmark & --
& 0.04/1.44 \\ \midrule
DSAC\cite{DSAC} & SCR & + & + &  &  & 0.20/6.30 & F \\
DSAC++\cite{DSAC++} & SCR & + & + &  &  & 0.08/2.40 & F \\
DSAC*\cite{brachmann2021visual} & SCR & + & + &  &  & 0.03/1.40 & F \\ \midrule
PoseNet\cite{kendall2015posenet} & APR & + & + &  &  & 0.44/10.4 & F \\
BayesianPN\cite{kendall2016modelling} & APR & + & + &  &  & 0.47/9.81 & F \\
LSTM-PN\cite{walch2017image} & APR & + & + &  &  & 0.31/9.85 & F \\
IRPNet\cite{shavitferensirpnet} & APR & + & + &  &  & 0.23/8.49 & F \\
PoseNetLearn\cite{kendall2017geometric} & APR & + & + &  &  & 0.24/7.87 & F
\\
GeoPoseNet\cite{kendall2017geometric} & APR & + & + &  &  & 0.23/8.12 & F \\
Atloc\cite{wang2020atloc} & APR & + & + &  &  & 0.20/7.56 & F \\
MSPN\cite{blanton2020extending} & APR & + & + & \checkmark &  & 0.20m/8.41 &
F \\
MS-Transformer\cite{shavit2021learning} & APR & + & + & \checkmark &  &
0.18/7.28 & F \\ \midrule
NN-Net\cite{nn-net} & RPR & ++ & ++ & \checkmark & \checkmark & 0.21/9.30 &
2.31/13.0* \\
EssNet\cite{zhou2020learn} & RPR & ++ & ++ & \checkmark & \checkmark &
0.22/8.00 & 0.89/40.2* \\
NCEssNet\cite{zhou2020learn} & RPR & ++ & ++ & \checkmark & \checkmark &
0.21/7.50 & 0.82/26.2* \\
RelPoseGNN\cite{turkoglu2021visual} & RPR & ++ & ++ & \checkmark & \checkmark
& 0.16/5.20 & 0.36/13.6 \\
\textbf{Relformer (Ours)} & RPR & ++ & ++ & \checkmark & \checkmark &
0.18/6.27 & 0.34/9.72 \\ \bottomrule
\end{tabular}%
\end{table*}
\begin{table*}[tbh]
\caption{\textbf{Ablations of our network architecture on previously seen
scenes}. We report the median position and orientation errors (in meters and
degrees, respectively) for each scene and the average of medians across all
scenes. Performance is reported for a baseline architecture and when
extracting Feature Maps (FM), aggregating them with a Transformer Encoder
(TE) and when learning a 6D vector to recover the rotation matrix
through orthogonalization (6D).}
\label{tb:ablation-seen}\centering
\begin{tabular}{ccc|ccccccc|c}
\toprule
\textbf{FM} & \textbf{TE} & \textbf{6D} & \textbf{Chess} & \textbf{Fire} &
\textbf{Heads} & \textbf{Office} & \textbf{Pumpkin} & \textbf{Kitchen} &
\textbf{Stairs} & \textbf{Average} \\ \midrule
&  &  & 0.10/4.79 & 0.34/11.7 & 0.14/11.0 & 0.15/6.00 & 0.19/4.67 & 0.18/5.86
& 0.26/8.86 & 0.20/7.57 \\
\checkmark &  &  & 0.10/5.95 & 0.22/10.7 & 0.18/12.9 & 0.18/7.08 & 0.18/6.24
& 0.189/7.05 & 0.29/10.2 & 0.19/8.58 \\
\checkmark & \checkmark &  & 0.10/4.92 & 0.23/8.72 & 0.15/11.8 & 0.16/5.46 &
0.17/4.21 & 0.18/5.20 & 0.24/7.78 & 0.18/6.87 \\
\checkmark & \checkmark & \checkmark & 0.11/4.01 & 0.23/8.57 & 0.17/10.89 &
0.16/4.92 & 0.15/4.15 & 0.19/4.89 & 0.24/6.46 & 0.18/6.27 \\ \bottomrule
\end{tabular}%
\vspace{-6mm}
\end{table*}
\begin{table*}[tbh]
\caption{\textbf{Ablations of our network architecture on unseen scenes}. We
report the median position and orientation errors (in meters and degrees,
respectively) for each scene and the average of medians across all scenes,
when training with sets of six scenes and evaluating on the remaining scene.
Performance is reported for a baseline architecture and when extracting
Feature Maps (FM), aggregating them with a Transformer Encoder (TE) and when
learning a 6D vector to recover the rotation matrix through
orthogonalization (6D).}
\label{tb:ablation-unseen}\centering
\begin{tabular}{ccc|ccccccc|c}
\toprule
\textbf{FM} & \textbf{TE} & \textbf{6D} & \textbf{Chess} & \textbf{Fire} &
\textbf{Heads} & \textbf{Office} & \textbf{Pumpkin} & \textbf{Kitchen} &
\textbf{Stairs} & \textbf{Average} \\ \midrule
&  &  & 0.30/11.8 & 0.36/15.5 & 0.27/15.25 & 0.35/10.8 & 0.37/11.8 &
0.34/11.8 & 0.45/19.9 & 0.35/13.8 \\
\checkmark &  &  & 0.32/10.2 & 0.37/15.1 & 0.22/13.6 & 0.33/10.3 & 0.38/12.9
& 0.36/11.7 & 0.40/13.0 & 0.34 /12.4 \\
\checkmark & \checkmark &  & 0.27/8.80 & 0.35/12.0 & 0.25/12.6 & 0.30/7.94 &
0.36/8.17 & 0.32/8.36 & 0.42/10.1 & 0.32/9.71 \\
\checkmark & \checkmark & \checkmark & 0.27/8.50 & 0.34/10.4 & 0.28/12.1 &
0.32/8.24 & 0.40/9.06 & 0.32/9.23 & 0.425/10.5 & 0.34/9.72 \\ \bottomrule
\end{tabular}%
\end{table*}
\begin{table*}[tbh]
\caption{\textbf{Ablations of the backbone by localizing previously seen and
unseen environments}. We report the median position and orientation errors
(in meters and degrees, respectively) for the Fire scene (7scenes dataset),
when the scene data is included in (Seen) or excluded from (Unseen) the
training data.}
\label{tb:ablation-backbone}\centering
\begin{tabular}{lccc}
\toprule
\textbf{Backbone} & \textbf{Dimension} & \textbf{Seen (Fire)} & \textbf{%
Unseen (Fire)} \\ \midrule
ResNet34 & $7\times7\times512$ & 0.32/14.2 & {0.39}/{15.2} \\
ResNet50 & $7\times7\times2048$ & 0.27/13.3 & 0.42/15.8 \\
EfficientNet-B0 & $7\times7\times320$ & 0.24/11.8 & 0.34/11.6 \\ \bottomrule
\end{tabular}%
\vspace{-6mm}
\end{table*}
\begin{table*}[tbh]
\caption{\textbf{Ablations of feature maps and aggregation operators on the
7Scenes dataset}. We report the average of median position and orientation
errors (in meters and degrees, respectively) when varying on the type of
aggregation (Agg.) and the dimensions of the feature maps.}
\label{tb:ablation-maps}\centering
\begin{tabular}{lcc}
\toprule
\textbf{Agg.} & \textbf{Feature Maps} & \textbf{Average} \\ \midrule
2D Conv & $7\times7\times320$ / $14\times14\times112$ & 0.19/8.15 \\
2D Conv & $14\times14\times112$ / $28\times28\times40$ & 0.19/8.59 \\
Transformer Encoder & $7\times7\times320$ / $14\times14\times112$ & 0.18/7.35
\\
Transformer Encoder & $14\times14\times112$ / $28\times28\times40$ &
0.18/6.87 \\ \bottomrule
\end{tabular}%
\end{table*}
\begin{table*}[tbh]
\caption{\textbf{Ablations of rotation representation (7Scenes)}. We compare
the performance when learning with different rotation representations
(Repr.): Quaternion (Quat.), 6D and 9D. We report the median position and
orientation errors (in meters and degrees, respectively) for each scene and
the average of error medians across all scenes. }
\label{tb:ablation-rot}\centering
\begin{tabular}{l|ccccccc|c}
\toprule
\textbf{Repr.} & \textbf{Chess} & \textbf{Fire} & \textbf{Heads} & \textbf{%
Office} & \textbf{Pumpkin} & \textbf{Kitchen} & \textbf{Stairs} & \textbf{%
Average} \\ \midrule
Quat. & 0.10/4.92 & 0.23/8.72 & 0.15/11.8 & 0.16/5.46 & 0.17/4.21 & 0.18/5.20
& 0.24/7.78 & 0.18/6.87 \\
9D \cite{levinson2020analysis} & 0.11/4.66 & 0.23/9.24 & 0.17/11.8 &
0.16/5.13 & 0.17/3.99 & 0.19/4.83 & 0.26/8.25 & 0.18/6.84 \\
6D \cite{zhou2019continuity} & 0.11/4.01 & 0.23/8.57 & 0.17/10.89 & 0.16/4.92
& 0.15/4.15 & 0.19/4.89 & 0.24/6.46 & 0.18/6.27 \\ \bottomrule
\end{tabular}%
\vspace{-4mm}
\end{table*}

\section{Experimental Results}

This work focuses on studying and improving the generalization of RPRs.
Hence, we evaluated our model's performance by localizing previously seen
and unseen indoor and outdoor contemporary benchmarks, used to evaluate
regression methods. We compare our results to recent state-of-the-art
relative pose regression methods (Tables \ref{tb:7scenes-seen}, \ref%
{tb:7scenes-unseen}, \ref{tb:cambridge}) and also provide a broader
comparative analysis, across different classes of localization (Table \ref%
{tb:cross}). We further carry out ablations in order to assess the
particular contribution of the main building blocks of our architecture to
the performance and generalization of our model (Table \ref{tb:ablation-seen}%
,\ref{tb:ablation-unseen}). In addition, we evaluate different design
choices: the convolutional backbone used (Table \ref{tb:ablation-backbone}),
the dimension of the feature maps (Table \ref{tb:ablation-maps}) and the
rotation parameterization (Table \ref{tb:ablation-rot}).\newline
\textbf{Datasets.} The 7Scenes dataset \cite{glocker2013real} describes a
small-scale indoor environment ($\sim 1-10m^{2}$), captured inside an office
building. It consists of seven scenes and presents various localization
challenges such as motion blur, reflections, occlusions and repetitive
textures. The CambridgeLandmarks dataset \cite{kendall2015posenet} is a
mid-scale dataset ($\sim 900-5500m^{2}$), captured in an urban outdoor
environment. The dataset presents challenges that are typical to outdoor
localization, such as changing lighting conditions and variations in scale,
view points and trajectory. We evaluated our model in four (of six) scenes,
which are commonly used as benchmarks by regression methods.\newline
\textbf{Training Details.} We follow the training procedure of \cite%
{ShavitFerensIccv21} and apply the augmentation method suggested by \cite%
{kendall2015posenet}. Specifically, during training, images are first
rescaled such that their smaller edge is resized to $256$ pixels. We extract
random $224\times 224$ crops and randomly jitter the brightness, saturation,
and contrast. At test time, images are rescaled, and the center crop is used
for inference, with no further augmentations. Our model is optimized using
Adam, with $\beta _{1}=0.9$, $\beta _{2}=0.999$ and $\epsilon =10^{-10}$ and
a batch size of $8$. We used an initial learning rate of $\lambda =10^{-4}$
and a weight decay of $10^{-4}$ and trained for 30 (600) epochs for indoor
(outdoor) localization. Our loss is initialized as in \cite%
{kendall2017geometric}. The training pairs for the 7Scenes dataset are
generated as in \cite{ding2019camnet}. We further extend this dataset with
pairs from \cite{nn-net}. Pairs from the CambridgeLandmarks dataset are
computed by encoding images with a pre-trained NetVLAD model \cite%
{arandjelovic2016netvlad}, and fetching the nearest 40 neighbors based on
cosine similarity. At train time, we sample from the pool of neighbors to
generate different pairs. We provide the training pairs, code, configuration
files, and pre-trained models in order to support the reproducibility of our
results. All models were trained on a single NVIDIA RTX A5000 24GB GPU.
\newline
\textbf{Implementation Details.} We use EfficientNet-B0 \cite%
{pmlr-v97-tan19a} as our convolutional backbone. We extract the feature maps
$\mathbf{F}_{\mathbf{trans}}$ and $\mathbf{F}_{\mathbf{rot}}$ at resolutions
of $14\times 14\times 112$ and $28\times 28\times 40$, respectively, as
these endpoints were previously shown to provide useful features for camera
pose regression \cite{shavit2021learning} (we reconfirm this choice in our
ablation study). We set $C_{h}=512$ for the dimension of our projected
paired feature maps and their respective positional encoding. Our
Transformer Encoder is implemented with a standard architecture consisting
of six layers and a dropout of $p=0.1$. Each layer contains a MHA module
with eight heads and an MLP which expands the input dimension (512) to 2048
with a hidden layer and gelu non-linearity and then decrease it back to 512
with another FC layer. Our MLP regressor heads preserve the dimension of the
input and then regress the target with an FC layer. Evaluation on the
7Scenes dataset is done using the pairs (query and reference image) from
\cite{nn-net}, in line with \cite{turkoglu2021visual}. For the
CambridgeLandmarks dataset, for each query, we fetch the nearest neighbor
using a pre-trained NetVLAD model \cite{arandjelovic2016netvlad}.
\subsection{Localization in Seen and Unseen Scenes}
\vspace{-2mm}
Tables \ref{tb:7scenes-seen},\ref{tb:7scenes-unseen} compare the median
position and orientation errors of our model (Relformer) and recent
state-of-the-art RPRs for localizing the 7Scenes dataset. We consider
localization in seen scenes, where the training data contain images from all
seven scenes (Table \ref{tb:7scenes-seen}) and localization in unseen scenes
(Table \ref{tb:7scenes-unseen}), where each time we train on six scenes and
report the pose error for the remaining unseen scene. Relformer consistently
outperforms current state-of-the-art RPRs when localizing in unseen scenes,
while maintaining competitive performance for seen scenes. In particular, we
note a significant improvement in regressing the rotation in unseen scenes.
Table \ref{tb:cambridge} reports the average of median position and
orientation errors for the CambridgeLandmarks dataset. We observe a similar
trend, where our model achieves a significant improvement in unseen scenes,
while performing similarly to other methods in seen scenes. We also consider
a broader comparative analysis in order to examine the different tradeoffs
exhibited by different localization approaches. Table \ref{tb:cross}
compares representative methods from different localization families: IR,
SLP, SCR, APR and RPR. Methods in the SLP and SCR families achieve
state-of-the-art localization. The SLP approach enables accurate
localization in unseen scenes, but is slow and requires large storage. SCR
methods, on the other hand, are fast and light but need to be trained per
scene and fail to generalize in unseen scenes. Regression methods are less
accurate than the SOTA SLP and SCR approaches. APRs and RPRs perform
similarly on seen scenes. APRs are faster than RPRs and do not require extra
storage, but, as opposed to RPRs, they cannot localize in unseen scenes.
Hence, the key upside of RPRs over APRs is in their ability to generalize
to unseen scenes. Within the family of localization-by-regression (absolute and relative pose regression), our model localizes better in unseen scenes while outperforming APR approaches and maintaining competitive results compared to contemporary RPRs.

\subsection{Ablation Study}

We study the particular contributions of the main algorithmic components of
our model, differentiating it from contemporary RPRs (see Fig. \ref{fig:rpr}%
): feature maps (FM), paired aggregation with a Transformer Encoder (TE) and
a 6D rotation representation as a regression target (6D). Tables \ref%
{tb:ablation-seen} and \ref{tb:ablation-unseen} report position and
orientation errors when localizing in seen and unseen scenes from the
7Scenes dataset, respectively. Our baseline architecture (first row in
Tables \ref{tb:ablation-seen} and \ref{tb:ablation-unseen}), extracts global
descriptors instead of feature maps (1280-dimensional vectors), concatenates
them and passes them to two dedicated MLP heads to regress the translation
and rotation parameters (with the same architecture as the MLP heads used in
our fully proposed model). We also compare our full architecture to a
configuration where feature maps are aggregated with two 2D convolution
layers and relu non-linearity after concatenation and projection (second row
in Tables \ref{tb:ablation-seen} and \ref{tb:ablation-unseen}). Our full
model and the latter two configurations are trained using the Loss in Eq. %
\ref{equ:learnable pose loss} with the quaternion representation. We compare
their performance to our full Relformer configuration, which uses the 6D
rotation parameterization. The use of a Transformer Encoder to aggregate
feature maps and the adoption of 6D parameterization have resulted in a
notable improvement in orientation estimation, as demonstrated in Table %
\ref{tb:ablation-seen}. The use of feature maps instead of a global
descriptor computed through average pooling, together with the proposed
paired aggregation, leads to improved position estimation. A similar trend
is observed when localizing in unseen scenes, and such algorithmic
components can be adopted by other relative pose regression methods to
improve their generalization. Table \ref{tb:ablation-backbone} further studies the choice of the backbone
used to extract feature maps. We report the performance for the Fire scene,
when aggregating the features with a 2D convolution (the same configuration
as in the second row of Tables \ref{tb:ablation-seen} and \ref%
{tb:ablation-unseen}). To allow for a fair comparison, we consider
feature maps with the same height and width. The Efficient-B0 backbone
achieves the best position and orientation errors for both seen and unseen
scenarios. Interestingly, we observe that features from the same backbone
type (ResNet\cite{he2016deep}) with a higher dimension yield better
performance when the scene is seen during training but degrade the
generalization of the model, compared to more compact features of the same
resolution (ResNet50 versus ResNet34, respectively). Finally, we also validate two main aspects of our architecture design: the
aggregation operation on different dimensions of the extracted feature maps
(Table \ref{tb:ablation-maps}) and the parameterization of the rotation
(Table \ref{tb:ablation-rot}). Consistent with our other experiments, we
observe that the aggregation of flattened paired feature maps, while
attending to nearby and long-range correlations (Transformer Encoder) yields
better results, compared to local aggregation with 2D convolutions (Table %
\ref{tb:ablation-maps}). This trend is consistent for different feature
maps. Extracting the $14\times 14$ and $28\times 28$ resolution for learning
the translation- and rotation-related features with Transformer Encoders
achieves better performance compared to the other configuration tested. When comparing models trained with different rotation parameterizations
(Table \ref{tb:ablation-rot}), we find that the 6D representation achieves
the lowest orientation error on average, with a consistent improvement of $%
\sim 0.5^{\circ }$ between scenes. We note that the continuous
parameterizations tested (6D and 9D) yield better results compared to representation by quaternions, which is consistent with the findings of \cite%
{zhou2019continuity}.
\subsection{Limitations and Future Work}
\vspace{-1mm}
We note that, while our model outperforms other regression methods in unseen
scenes, its generalization differs greatly for position and orientation. For
example, when examining the orientation errors for unseen scenes in the
7Scenes dataset, our model not only significantly improves the performance
compared to other methods, but also achieves an error that is similar in
magnitude to the error achieved in previously seen scenes. For position
estimation, however, the improvement is minor and the degradation between
seen and unseen scenes is larger. We attribute that to the different visual
cues required to predict translation and rotation. Rotation estimation is
more scale-invariant and can rely on cues such as lines and shadows even
without significant visual overlap \cite{Cai2021Extreme}. However, the
estimation of translations requires some knowledge of the scene scale.
Whether this can be mitigated through further research or is an inherent
limitation of RPRs, is a topic for future research. In addition, we note
that our method requires larger storage compared to current relative pose
regression methods, since feature maps, instead of global descriptors, need
to be saved. This limitation can potentially be mitigated by compression and
reconstruction.

\section{Conclusions}

We presented a novel approach for relative pose regression, which pairs and
aggregates feature maps with Transformer Encoders. Our model surpasses
current regression methods in unseen scenes while maintaining competitive
localization in seen scenes. Through multiple ablations across indoor and
outdoor benchmarks, we show that attention-based aggregation of paired
feature maps is key for improving the localization of RPRs. In addition, we
take advantage of recent advances in rotation representation and learn a
modified parameterization of the camera pose, which reduces the orientation
errors of our model. Our proposed approach is modular and can be adopted by
RPRs in order to improve their performance and generalization.

{\small
\bibliographystyle{ieee_fullname}
\bibliography{localization}
}

\end{document}